%% file: main.tex
\documentclass[10pt,twocolumn,letterpaper]{article}
\usepackage[pagenumbers]{cvpr}
\usepackage[dvipsnames]{xcolor}

\input{preamble}
\definecolor{cvprblue}{rgb}{0.21,0.49,0.74}
\usepackage[pagebackref,breaklinks,colorlinks,citecolor=cvprblue]{hyperref}
\usepackage[accsupp]{axessibility}

\usepackage[capitalize]{cleveref}
\crefname{section}{Sec.}{Secs.}
\Crefname{section}{Section}{Sections}
\Crefname{table}{Table}{Tables}
\crefname{table}{Tab.}{Tabs.}

\def\paperID{XXXX}
\def\confName{CVPR}
\def\confYear{2024}

\title{\vspace{-0.6cm}\textit{SketchINR}: A First Look into Sketches as Implicit Neural Representations\\[-0.6cm]}

\author{\href{https://hmrishavbandy.github.io/}{Hmrishav Bandyopadhyay}\textsuperscript{1} \hspace{.2cm} \href{https://ayankumarbhunia.github.io/}{Ayan Kumar Bhunia}\textsuperscript{1} \hspace{.2cm}  \href{http://www.pinakinathc.me/}{Pinaki Nath Chowdhury}\textsuperscript{1} \hspace{.2cm} \href{https://aneeshan95.github.io/}{Aneeshan Sain}\textsuperscript{1} \\ \href{https://www.surrey.ac.uk/people/tao-xiang}{Tao Xiang}\textsuperscript{1,2} \hspace{.2cm}  \href{https://homepages.inf.ed.ac.uk/thospeda/}{Timothy Hospedales} \textsuperscript{3}\hspace{.2cm} \href{https://personalpages.surrey.ac.uk/y.song/}{Yi-Zhe Song}\textsuperscript{1,2} \\
\textsuperscript{1}SketchX, CVSSP, University of Surrey, United Kingdom.  \\
\textsuperscript{2}iFlyTek-Surrey Joint Research Centre on Artificial Intelligence.\\
\textsuperscript{3}University of Edinburgh, United Kingdom\\
{\tt\small \{h.bandyopadhyay, a.bhunia, p.chowdhury, a.sain, t.xiang, y.song\}@surrey.ac.uk} \\ {\tt\small  t.hospedales@ed.ac.uk \vspace{-0.5cm}}
}

\newcommand\myfigure{
\centering
\vspace{-0.4cm}
\captionsetup{type=figure} 
\includegraphics[trim={0 0 1.5cm 0}, width=\textwidth]{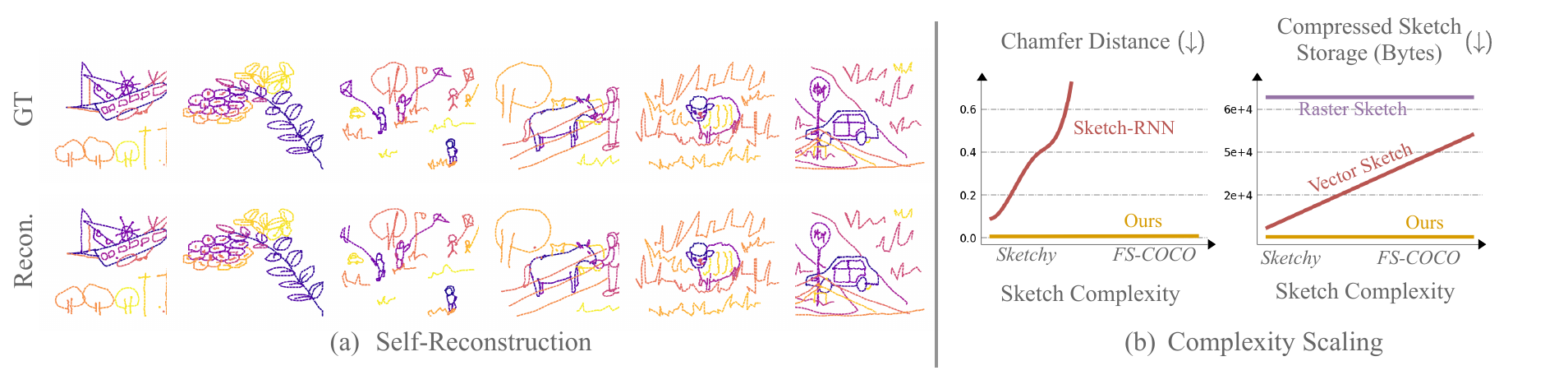} \\[-0.4cm]   
    \caption{SketchINR is an implicit neural representation for sequential vector sketches. It is the first neural representation with sufficient fidelity to be a drop-in replacement for the raw sketch data (a). The representation is significantly higher fidelity than existing learned representations such as SketchRNN \cite{ha2017neural}, especially for more complex sketches (b, left). SketchINR also provides state-of-the-art sketch compression, with a substantially more compact representation than either vector or raster sketches (b, right). \label{fig:teaser}}
    \vspace{0.3cm}

}
    
\begin{document}
\maketitle

\begin{abstract}
\vspace{-0.3cm}

We propose SketchINR, to advance the representation of vector sketches with implicit neural models. A variable length vector sketch is compressed into a latent space of fixed dimension that implicitly encodes the underlying shape as a function of time and strokes. The learned function predicts the $xy$ point coordinates in a sketch at each time and stroke. Despite its simplicity, SketchINR outperforms existing representations at multiple tasks: (i) Encoding an entire sketch dataset into a fixed size latent vector, SketchINR gives $60\times$ and $10\times$ data compression over raster and vector sketches, respectively. (ii) SketchINR's auto-decoder provides a much higher-fidelity representation than other learned vector sketch representations, and is uniquely able to scale to complex vector sketches such as \textit{FS-COCO}. (iii) SketchINR supports parallelisation that can decode/render $\sim$$100\times$ faster than other learned vector representations such as SketchRNN. (iv) SketchINR, for the first time, emulates the human ability to reproduce a sketch with varying abstraction in terms of number and complexity of strokes. As a first look at implicit sketches, SketchINR's  compact high-fidelity representation will support future work in modelling long and complex sketches.
\end{abstract}

\vspace{-0.6cm}
\section{Introduction}
\label{sec:intro}
\vspace{-0.1cm}
The prevalence of touch-screen devices has triggered significant research progress on sketches \cite{pixelor, vinker2022clipasso, binninger2024sens, ribeiro2020sketchformer, bandyopadhyay2024xai}. Modelling digital sketches has become an important challenge for learning systems that aim to stimulate human creativity \cite{das2021sketchode, creative-sketch}. Large-scale datasets \cite{ha2017neural, sangkloy2016sketchy, eitz2012humans} motivated several downstream applications like image retrieval \cite{collomosse2019livesketch, xu2018sketchmate}, image generation \cite{controlNet, picturethatsketch}, image editing \cite{aksan2018deepwriting, liu2021deflocnet}, and 3D content creation \cite{binninger2024sens, li2022free2cad, bandyopadhyay2024doodle}, among others \cite{vinker2023clipascene, xing2023diffsketcher}. 

These applications are underpinned by sketches captured in raw form either as raster or vector images, and usually encoded into derived representations by ConvNets \cite{collomosse2019livesketch}, splines, and so on. For over a decade, discourse around what is a “good” representation of human sketches has persisted \cite{hertzmann2020line}. A now substantial body of work has focused on representations for sequential vector sketches that model both explicit strokes, and their drawing over time — most famously by the auto-regressive SketchRNN \cite{ha2017neural}, but also using representations such as parametric curves \cite{das2020beziersketch}, Gaussians \cite{aksan2020cose}, and Neural ODEs \cite{das2021sketchode}.  We introduce a novel continuous-time representation of sequential vector sketches by taking an implicit neural representation perspective on sketches for the first time.

First, we review the limitations of current sketch representations. The two most popular raw sketch representations are (i) raster sketch -- a black and white image of a line drawing, and (ii) vector sketch -- a temporal sequence of points and strokes. Raster sketch has a large but fixed storage (\eg, $256 \times 256$). Its compatibility with ConvNets \cite{yu2016sketch} made raster sketches popular but they lack the temporal information (i.e., stroke order) necessary for generative modelling \cite{das2021sketchode}. Vector sketches efficiently store the $xy$-coordinates of points and strokes ($N \times 2$). However, the storage increases with the length of sketch $N$ ( \cref{fig:teaser}b). (iii) Raw vector sketches can be encoded by popular learned sketch representations such as SketchRNN \cite{ha2017neural} and others \cite{das2020beziersketch,sketchBert,das2021sketchode}. However, for longer ($N \geq 1000$) scene sketches, the quadratic computational cost for transformers becomes intractable \cite{sketchBert} and recurrent nets become unstable \cite{ha2017neural}. Nevertheless, most of these still suffer from slow sequential autoregressive decoding, and they mostly cannot accurately reconstruct complex sketches with many strokes. To summarise, current sketch representations either discard temporal information (raster), scale poorly in storage size (vector), or become slow and inaccurate with complex sketches (learned autoregressive vector). 

To overcome the above issues, we propose an implicit neural representation for sketches -- SketchINR. We piggyback on the well-established \cite{binninger2024sens, hertz2022spaghetti, park2019deepsdf, shap-e} potential of implicit functions \cite{implicit-function} to encode an underlying shape as a function of time/location. Different to standard implicits \cite{park2019deepsdf, mescheder2019occupancy}, the input of sketches as parametric functions is a hierarchy of stroke sequence and point sequence. We develop SketchINR to generate points ($xy$-coordinates) given both time and stroke inputs. To represent multiple sketches, SketchINR learns a fixed size latent space and a conditional implicit function that generates $xy$-coordinates of any sketch instance given its corresponding latent code.

Despite its simplicity, SketchINR is competitive with existing sketch representations. (i) It can encode an entire sketch dataset into a compact latent space (\eg, $\mathbb{R}^{N\times512}$) and decode using a fixed size function (\eg, an $8$ layer MLP). Compared to raster sketches ($256 \times 256$), SketchINR provides $128\times$ storage compression. While simple object-level vector sketches (length $\leq 300$) have comparable storage with SketchINR ($\mathbb{R}^{512}$), for practical scene-level \cite{chowdhury2022fs} vector sketches (length $\geq 1000$), SketchINR provides $10\times$ better compression (\cref{fig:teaser}b). (ii) For decoding/generation, SketchINR can parallelly  predict the $xy$-coordinates for all time and strokes.  
In practice, this is $100\times$ faster than an autoregressive vector sketch generator with length $\geq 300$. (iii) SketchINR can accurately represent complex sketches (\cref{fig:teaser}a,b) unlike competitors that suffer from complexity limitations \cite{das2021sketchode,das2020beziersketch} or the length limitations of auto-regression \cite{ha2017neural}. The encoding fidelity is sufficiently high that SketchINR provides a drop-in replacement for raw sketch data, while being more compact. (iv) SketchINR supports diverse applications spanning smooth latent space interpolation, and sketch generation, sketch completion. Uniquely, it also supports sketch abstraction -- the human-like ability to replicate the essence of a sketch with a variable number and complexity of strokes \cite{linePLOS,hertzmann2020line}. 

In summary, our contributions are: 
(i) We present the first implicit neural representation approach to vector sketch modelling, extending INRs as a function of time and strokes, and defining a training objective for learning them.
(ii) SketchINR provides the first learned representation capable of representing complex sketches compactly and with high-fidelity (\cref{fig:teaser}a). Based on our compact high-fidelity representation we introduce the task of sketch compression, and demonstrate excellent compression results (\cref{fig:teaser}b)). 
(iii) We further demonstrate SketchINR's applicability to a variety of sketch related tasks including generation, interpolation, completion and abstraction.

\section{Related Work}

\noindent \textbf{Sketch Representations:}
Digital sketches are predominantly captured as a function of `time' \cite{ha2017neural},  in a \textit{sequence} of coordinates traced by an artist on a canvas. Recent works \cite{ha2017neural,eitz2012humans} emphasise this temporal order of coordinates as an indicator of their relative importance in depicting a sketch-concept, as humans draw in a coarse-to-fine fashion. As such, they \cite{ha2017neural, aksan2020cose, das2021cloud2curve,bhunia2021vectorization} exploit temporal information in sketches for downstream tasks like sketch-assisted retrieval \cite{bhunia2021vectorization}, generation \cite{das2021cloud2curve,ha2017neural,aksan2020cose} and modelling \cite{li2022free2cad}. Vector sequences are further parameterised, for representation with parametric curves like B\'eziers \cite{das2020beziersketch,vinker2022clipasso} and B-Splines \cite{revow1996using, zheng2012fast} for early-handwritten digit representation \cite{revow1996using} to recent representations of complex structures like real-world objects \cite{das2020beziersketch,vinker2022clipasso} and scenes \cite{vinker2023clipascene}. Discarding temporal significance of strokes, vector-sketches can be converted to raster images by rendering coordinate strokes \cite{bresenham1965algorithm} on a 2D canvas. While raster sketches are less informative than their vector counterparts \cite{ha2017neural}, they offer enhanced applicability by expanding beyond digital sketches to \textit{paper}-sketches \cite{das2021cloud2curve} where stroke order is already lost. Raster representations are encoded with translation-invariant networks like CNNs \cite{yu2015sketch} and Vision-Transformers \cite{binninger2024sens}, making them extremely useful to represent object-level information.

\noindent \textbf{Non-Parametric Modelling:}
Both vector sequences and raster renderings of these sequences are non-parametric representations \cite{das2020beziersketch} of sketches. To preserve temporal stroke order, vector sketches are handled as sequences \cite{ha2017neural} of coordinates rather than unordered sets. Hence, they are encoded with position-aware networks like RNNs \cite{ha2017neural}, LSTMs \cite{xu2018sketchmate, collomosse2019livesketch}, and Transformers \cite{ribeiro2020sketchformer,carlier2020deepsvg}. Without positional cues, raster sketches are processed by spatially-aware CNNs \cite{yu2015sketch} and Vision Transformers \cite{sain2023clip}. Semantic encodings of non-parametric representations capture the expressive nature of sketch, allowing its use as a creative input for downstream segmentation \cite{hu2020sketch}, retrieval \cite{bhunia2021vectorization}, and editing \cite{liu2021deflocnet}, as well as for interactive tasks like object-detection \cite{chowdhury2023can} and image-inpainting \cite{yu2019free}. These encodings are further used in auto-regressive \cite{ha2017neural, wang2021sketchembednet} and VAE-like \cite{carlier2020deepsvg} modelling of vector coordinate sequences for the generation of textual-characters \cite{aksan2018deepwriting} and object sketches \cite{ha2017neural}.

\noindent \textbf{Parametric Representation learning:} Parametric Splines \cite{de1978practical}, such as Beizer curves \cite{masood2010efficient}, approximate vector sketches by per-sample fitting on individual sketch samples. Amortised frameworks bypass this per-sample optimisation by inferring curve control points \cite{das2020beziersketch} and degree \cite{das2021cloud2curve} from individual stroke features. Recent works exploit the representative power of parametric curves for generation of simple characters \cite{ganin2018synthesizing} to complex object \cite{vinker2022clipasso, das2020beziersketch, das2021cloud2curve} and scene level sketches\cite{vinker2023clipascene}. Both non-parametric and curve-fitted B\'ezier parametric representations capture the \textit{explicit} form of a sketch as spatial and sequential forms. 

Beyond explicit representations of sketches, we learn visual implicits to capture spatial dynamics in vector-sketches. Learned implicits can be used for sketch reconstruction with particular control over abstraction through \textit{number} of reconstructed strokes. Somewhat similar to implicit representations for vector sketches, CoSE \cite{aksan2020cose} parameterises Gaussian Mixtures on individual strokes, learning one implicit for one stroke only. New implicits are generated auto-regressively from previous predictions for auto-regressive generation and completion of sketches. Unlike CoSE, we learn a single implicit for the entire sketch, thus having one implicit parameter to sample the sketch in one pass. While SketchODE \cite{das2021sketchode} similarly parameterises Neural Ordinary Differential Equations \cite{chen2018neural} to capture vector-sketch dynamics, it's optimisation is extremely slow (as much as $120\times$ slower) because of high time complexity in solving higher order differentials. Three dimensional implicits like Signed-Distance \cite{park2019deepsdf} and Occupancy \cite{mescheder2019occupancy} functions learned from 3D point-clouds are particularly analogous to our implicit representation.

\noindent \textbf{Sketch Abstraction:} The continuity of human cognition in visual perception \cite{koffka2013principles} is directly reflected in how we sketch. This enables versatility \cite{yu2016sketch} in sketch, allowing us to express a huge range of thoughts and visions directly on paper in the form of coarse-grained `ideas' \cite{hu2018sketch} and fine-grained `objects' \cite{yu2016sketch}. Sketch abstraction, as a direct inverse of this granularity, has been studied in depth as a function of (i) drawing time: humans draw most representative strokes first \cite{ha2017neural}, and (ii) compactness: in a constrained stroke setting, salient strokes are prioritised \cite{muhammad2018learning}. As abstraction gives sketches a human touch, generative modelling of sketches is focused on abstraction towards making generated drawings ``more humane" \cite{das2023chirodiff} and augmenting limited sketch datasets with simulated sketches from photos \cite{vinker2022clipasso,vinker2023clipascene}. Importantly, recent works \cite{vinker2022clipasso} model abstraction as a function of \textit{strokes} to explicitly control abstraction levels by restricting the number of generated strokes.
Here, we represent sketches as visual implicits which allows us to reconstruct them at varying levels of abstraction and detail by controlling the number of reconstructed strokes. We demonstrate several downstream applications of this representation, particularly sketch compression, generation and completion.

\section{Methodology}
\noindent \textbf{Overview:} 
Vector sketches are captured as pen movements on a digital canvas, represented by coordinates \((x,y) \in [0,1]\) with binary pen-states \(\delta \in \{0:\text{pen-down},1:\text{pen-up}\}\) at those coordinates. A sketch \(p = \{(x_i,y_i,\delta_i)\}_{i=1}^{N}\) comprises \(N\) vector \textit{way-points} (coordinates), and is portrayed (rendered) by retracing these points on a 2D canvas. This process, termed as rasterisation produces a pixelated raster sketch $p^{s}$ on the canvas. In this work, we leverage scalability \cite{park2019deepsdf} in implicit functions to learn a \textit{flexible} representation for vector sketches with raster guidance. 

\subsection{Learning Visual Implicits for Sketches} 

\begin{figure}[!h]
    \centering
    \includegraphics[trim={2cm 0 0 0}, width=\linewidth]{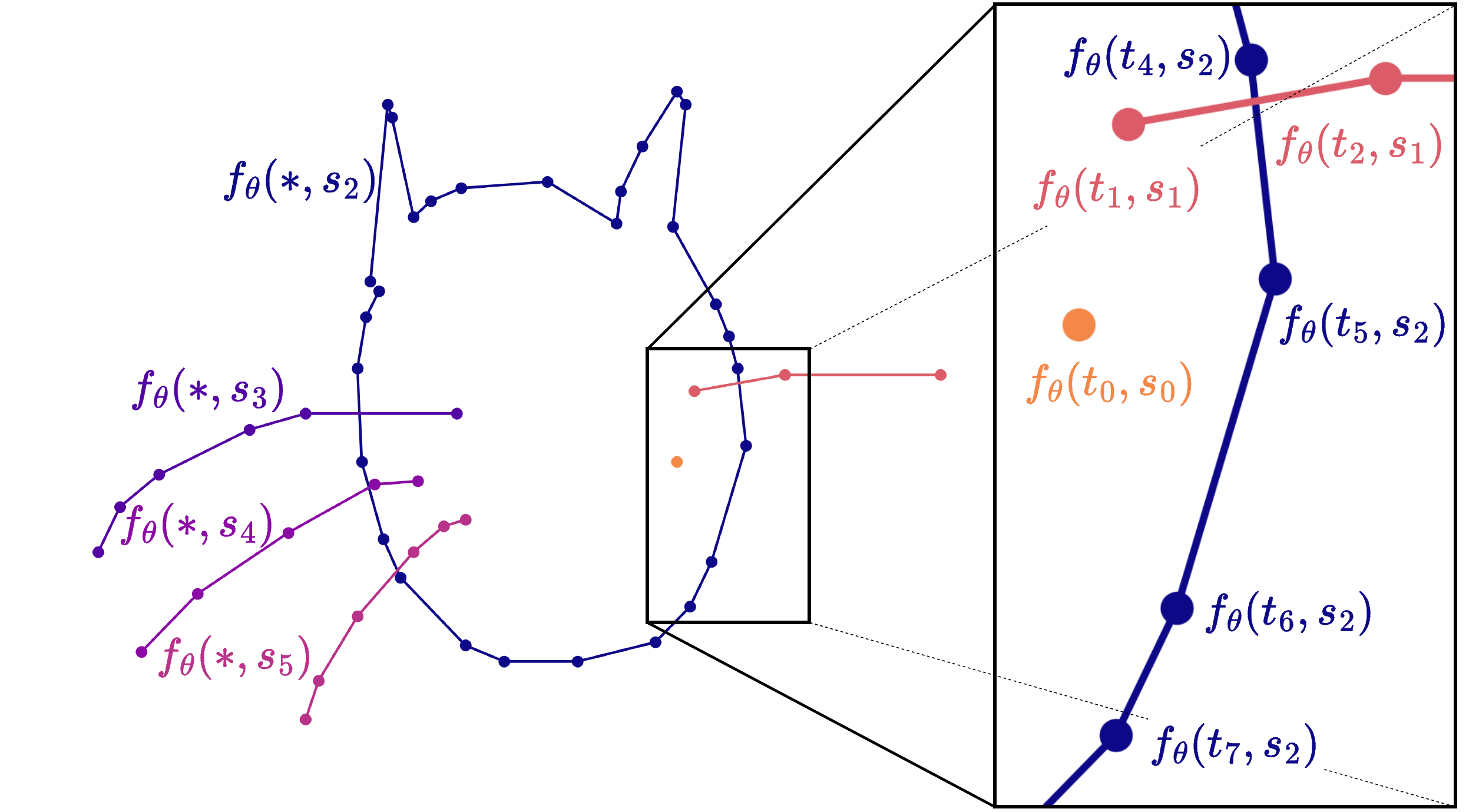}
    \caption{{We model sketch as a function $f_{\theta}(t_{j}, s_{k})$ of \(J\) timestamps and \(K\) strokes. Timestamps $\{t_a, \dots,t_b\}$ where $t_j\in [0,1)$ correspond to \textit{way-points} for a specific stroke $s_k$. Finally, $f_\theta$ with learned weights $\theta$ can be sampled with arbitrary number of strokes as $t_j\in\{0, \frac{1}{J}, \dots, \frac{J-1}{J}\}$  and $s_k \in \{0, \frac{1}{5}, \dots, \frac{4}{5}\}$ or $s_k \in \{0, \frac{1}{10}, \dots, \frac{9}{10}\}$, leading to increasing or decreasing abstraction for $K=5$ or $K=10$, respectively.}}
    \label{fig:sketch-implicits}
\end{figure}

\noindent Vector sketches can be modelled as implicits, where a sketch is represented as function of time \( f_{\theta}(t_{j}): \mathbb{R} \to \mathbb{R}^{2+1} \) parameterised by \(\theta\). Each time-stamp \(t_{j} \in [0, 1)\) is mapped to a pen position in the form of $xy$-coordinates \((x,y) \in \mathbb{R}^2\) and binary pen-states \(\delta \in \mathbb{R}^1\). The entire sketch can be reconstructed as a vector sequence by sampling \(t\) from \(0\) to \(1\) with a learned \(\theta\) as \(\mathbf{p} = \{f_{\theta}(0), \dots, f_{\theta}(t_{j}), \dots f_{\theta}(\frac{J-1}{J})\}\) for a total of \(J\) timestamps. This in itself is a flexible representation for a vector sketch, as it allows for sampling with arbitrary timestamps, thus potentially increasing or decreasing resolution of the resulting sketch by modulating number of strokes with sampling resolution (\(t : \{0,\frac{1}{3},\frac{2}{3}\} \text{ v/s } t:\{0,\frac{1}{10},\dots,\frac{9}{10}\}\)). For increased control over reconstruction granularity in the form of abstraction, we model \(f_\theta\) on an additional variable in the form of strokestamps \(\{s_k\}_{k=1}^{K}\), representing the sketch as \(\hat{p} = \{f_{\theta}(0,0), \dots, f_{\theta}(t_{j},s_{k}), \dots, f_\theta(\frac{J-1}{J},\frac{K-1}{K})\}\) for \(K\) strokes (\cref{fig:sketch-implicits}). These strokestamps provide explicit control over pen-up and pen-down states, allowing us to reconstruct the sketch with re-defined granularity during inference (\eg \(7\) or \(9\) strokes). Specifically, strokestamps represent the current stroke number as a fraction of the total number of strokes \(K\) which allows us to change strokes by changing \(s_k\). Pen-states \(\delta\) are thus explicitly determined from strokestamps only by toggling when \(s_k\) changes value. 
Our practical implementation of sketch implicits follows:
\begin{equation}\label{eq: sketch-implicit}
    f_{\theta}(t_{j}, \Delta t, s_{k}, \Delta s) = \mathrm{p}_{j}
\end{equation}

where, time instances are computed as \(t_{j} = t_{j-1} + \Delta t\), \(t_{0} = 0\), and \(\Delta t = \frac{1}{J}\). Similarly, $K$-stroke instances are computed as $s_{k} = s_{k-1} + \Delta s$, $s_{0} = 0$, and $\Delta s = \frac{1}{K}$. To exploit bias in neural networks towards learning low frequency functions \cite{mildenhall2021nerf,rahaman2019spectral}, we smoothen out high frequency variations \((0 \text{to} \frac{1}{K},0 \text{to} \frac{1}{S})\) for timestamps and strokestamps by mapping them from \(\mathbb{R}\) to a higher dimension \( \mathbb{R}^{2L}\) using positional embeddings $\text{PE}(\cdot)$ as
\begin{equation}
    \begin{split}
        \text{PE}(x) = & [\sin(10^{-4/L} x), \cos(10^{-4/L} x), \\
        & \dots, \sin(10^{-4L/L} x), \cos(10^{-4L/L} x)]
    \end{split}
\end{equation}
for $x \in \{t_{n}, \Delta t, s_{k}, \Delta s\}$. We uniformly distribute $J$ timestamps across $K$ strokes, where each stroke has $J/K$ points. 
For data augmentation, we train $f_{\theta}$ with varying $J$ and $K$ ($\pm 50\%$ of ground truth value). This helps $f_{\theta}$ to scale to different levels of abstraction, and number of points and strokes during inference.

\subsection{Loss functions}

To fit our implicit function \(f_\theta\) to a particular sketch instance \(p = \{(x_0,y_0,\delta_0),\dots,(x_{M-1},y_{M-1},\delta_{M-1})\}\) with \(M\) vector way-points \(\{p_m\}_{m=0}^{M-1}\), a naive approach would be to optimise \(\theta\) with mean squared error against the ground truth vectors as:
\begin{equation}\label{eq: mse}
    \mathcal{L}_{\text{MSE}} = \sum_{m=0}^{M}\Big|\Big|f_\theta\Big(\frac{m}{M},\frac{k}{K}\Big) - p_m\Big|\Big|_2
\end{equation}
where each predicted coordinate and pen-state is optimised directly against corresponding vector way-points from the ground truth. However, this rigid point-to-point matching enforces memorisation of particular points and strokes without any spatial understanding. As such, sampling with a different resolution at this stage yields noisy output that does not correspond to the raster sketch. To explicitly optimise for sampling flexibility, we introduce a visual loss, penalising sketch reconstructions that do not match visually. For this, we sample a template grid \(G\) of 2D coordinates and compute the region of influence of each vector-waypoint \(p_m\) of stroke \(s_k\) on coordinate \(g\) as an intensity map \cite{alaniz2022abstracting, wang2021sketchembednet}:
\begin{equation}\label{eq: dist_field}
    \mathrm{I}_{k}(g, s_{k}) = \hspace{-0.3cm} \max_{p_{m} \in s_{k}, r \in [0, 1]} \hspace{-0.3cm} \exp(\gamma ||g - rp_{m} - (1-r) p_{m+1}||_{2})
\end{equation}
where intensity \(\mathrm{I}_k\) is smoothed with \(\gamma\) for points \(p_m\) in stroke \(s_k\) where $m \in \{0, \dots, \frac{K-1}{K}\}$. The intensity map for the entire sketch is formulated as a summation of maps from all strokes for \(k\in(0,1]\) for both the ground truth sketch vector \(p\) with \(K\) strokes and the arbitrarily sampled sketch \(\hat{p}\) with \(\hat{K}\) strokes as \(\hat{\mathrm{I}}_k\), composing the loss as:
\begin{equation}
    \begin{split}
    \mathcal{L}_{V} = \Big|\Big| \sum_{k=0}^{\frac{K-1}{K}} \mathrm{I}_{k} -\sum_{k=0}^{\frac{\hat{K}-1}{\hat{K}}} \mathrm{\hat{I}}_{k} \Big|\Big|_{2}
    \end{split}
\end{equation}

\begin{figure}[!h]
    \centering
    \includegraphics[trim={0cm 0 1cm 0}, width=\linewidth]{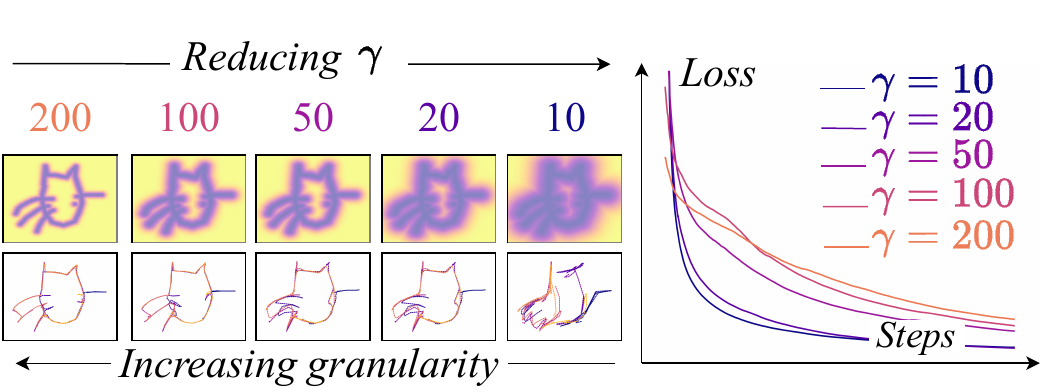}
    \vspace{-0.6cm}
    \caption{Effect of smoothing factor $\gamma$ on training: Reducing $\gamma$ leads to higher intensity in the surrounding region near stroke $s_{k}$. This is similar to stroke dilation of intensity map $\mathrm{I}_{k}$. A \textcolor{gamma_low}{lower} $\gamma$ leads to stable training (plot on right) but lacks fine-grained details. A \textcolor{gamma_high}{higher} $\gamma$ gives a fine-grained sketch but is harder to train.}
    \label{fig: effect-gamma}
    \vspace{-0cm}
\end{figure}

\noindent \textbf{Intensity Smoothing with $\mathbf{\gamma}$:} The sketch intensity map \(\mathrm{I}_{k}\) helps us to ground vector sketches to raster definites (\cref{eq: dist_field}) by a visual rendering of individual strokes where intensity in coordinates \(g \in G\) drops exponentially with distance from \(p_m \in s_k\). This exponential drop is controlled by a smoothing factor $\gamma$ as a hyperparameter, where (i) a low $\gamma$ yields thicker lines on $\mathrm{I}_{k}$ (low-quality sketch reconstructions) but offers better optimisation, while (ii) a high $\gamma$ yields more refined lines (higher quality reconstructions), making optimisation significantly slower (\cref{fig: effect-gamma}). 

\begin{figure*}
    \centering
    \includegraphics[trim={2cm 0 3cm 0}, width=\linewidth]{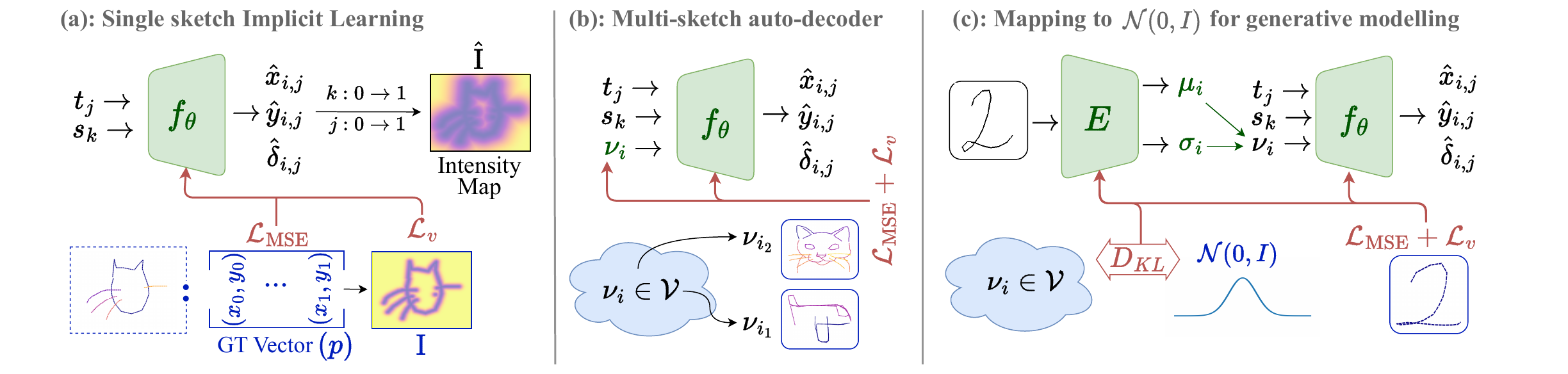}
    \vspace{-0.7cm}
    \caption{SketchINR model diagram. \textit{(a)} Embedding a single vector sketch as an implicit model $F$ mapping stroke and time $(\mathbf{s},\mathbf{t})$ to ink coordinate $\mathbf{p}$. \textit{(b)} Embedding a vector sketch dataset as a shared decoder and set of latent vectors $\mathcal{V}$. \textit{(c)} Training a generative model for implicit sketches by generating latent codes $\nu$ using an encoder $E$  that inputs raster sketches.\\[-0.7cm]}
    \label{fig:autodecoder-INR}
\end{figure*}

For faster optimisation, we assist the visual penalty with a weighted MSE loss from \cref{eq: mse} and a lower gamma, empirically set to \(\gamma = 150\). The loss function weighted with \(\lambda_\text{MSE}\), looks can be written as: 
\begin{equation}
    \mathcal{L}_{\text{implicit}} = \mathcal{L}_{V}^{\gamma=150} + \lambda_{\text{MSE}} \cdot \mathcal{L_\text{MSE}}
    \label{equation: all}
\end{equation}
where we set \(\lambda_\text{MSE}\) to \(0.7\). This, however, inherently introduces bias in our optimisation to match the order of strokes with the ground truth order to some extent. Future works could ideally set \(\lambda_\text{MSE}=0\) if they obtain faster convergence from a better engineered implicit function \(f_\theta\). \cref{fig:autodecoder-INR}a summarises our implicit sketch learning framework.

\vspace{-0.3cm}
\subsection{Generalising to Multiple Sketches}\label{sec:multi}
To efficiently represent multiple sketch implicits with a single global and general function, we modify \cref{eq: sketch-implicit} to include a sketch descriptor \(\nu_i \in \mathbb{R}^{d}\) for each sketch \(i\). The function \(f_\theta(\nu_i, t, \Delta t, s_k, \Delta s)\) is approximated with a fully connected auto-decoder network, similar to DeepSDF \cite{park2019deepsdf}. Thus, feature descriptors \(\nu_i\) are hidden vectors, learned jointly with the function parameters \(\theta\) on a training dataset (\(i \in D\)). Similar to single sketch optimisation, we train on multiple samples with a combination of MSE and visual loss (\cref{equation: all}). \cref{fig:autodecoder-INR}b summarises our multi-sketch auto-decoder through block diagrams.

\noindent \textbf{Generative Modelling:} The generalised auto-decoder, enables simple generative modelling of sketches by generating their feature descriptors \(\nu_i \in \mathcal{V}\), which allows us to create novel sketch implicits. Towards this, we train a Variational Auto-Encoder \cite{kingma2013auto} as our generator, where we replace the decoder with our pre-trained implicit decoder. For variational inference, our encoder \(E(\cdot)\) represents each raster sketch instance $i$ as a set of \(d\) Gaussians with mean \(\mu_i \in \mathbb{R}^d\) and variance \(\sigma_i \in \mathbb{R}^d\) which are re-parameterised \cite{kingma2013auto} to a latent encoding as \(z_i = \mu_i + \sigma_i \cdot \epsilon\) where \(\epsilon \in \mathcal{N}(\textbf{0},\textbf{I})\). This latent is then projected to the \(d\)-dimensional implicit latent space \(\nu_i \in \mathcal{V}\). We train the encoder as a ResNet-18 network with (i) reconstruction loss from both our visual and vector spaces (\cref{equation: all}) and (ii) KL-divergence loss to bring the distribution closer to a unit normal \(\mathcal{N}(\textbf{0},\textbf{I})\). The loss function for the encoder can be written as: 
\begin{equation}
    \mathcal{L}_{\text{enc}} = \mathcal{L}_{\text{implicit}} - \beta \cdot D_{KL}(q(z_i|i),\mathcal{N}(\textbf{0},\textbf{I}))
    \label{eq: enc}
\end{equation}
where \(\beta\) is a weighing factor, set to \(0.7\) and \(q(z_i|i)\) represents the underlying probability distribution modelled with encoder \(E\) that infers \(z_i\) for a given sketch \(i\). \cref{fig:autodecoder-INR}c summarises our implicit sketch generative model. The decoder can be used for sketch generation, while the encoder can encode raster sketches for vectorisation.

\section{Applications}

\noindent \textbf{Datasets:} We learn implicits for sketches of various complexities and abstraction levels, ranging from highly complex scene sketches that correspond to a photo scenery \cite{chowdhury2022fs} to abstract doodles \cite{ha2017neural} drawn from memory. Based off complexity in downstream applications, we use scene sketches to demonstrate self-reconstruction and sketch compression, while focusing on simpler object level sketches for more complex tasks like sketch mixing (interpolation) and generation for faster optimisation. Specifically, we use the \textit{FS-COCO} \cite{chowdhury2022fs}, \textit{Sketchy} \cite{sangkloy2016sketchy}, and \textit{Quick-Draw!} \cite{ha2017neural} datasets. (i) \textit{FS-COCO } \cite{chowdhury2022fs} contains 10,000 hand-drawn sketches of complex MS-COCO \cite{lin2014microsoft} scenes, drawn by amateurs without time constraints. Sketches in \textit{FS-COCO} have on a median of 64 strokes, resulting in the sketch having \(\sim\)\(3000\) waypoints. (ii) Comparatively simpler than scene sketches, \textit{Sketchy} \cite{sangkloy2016sketchy} consists of \(\sim\)75K sketches of 12.5K object photographs from 125 categories. These sketches similar to \textit{FS-COCO} are drawn from reference photographs and have photo-sketch pairs. (iii) \textit{Quick-Draw!} \cite{ha2017neural} sketches are drawn from memory and in a constrained time settings ( \(<20\)s). Hence, these sketches (doodles) are abstract in nature, lacking any pre-defined configuration or orientation from reference photos. In addition to these three datasets, we use a vectorised form of the MNIST dataset, \textit{Vector-MNIST} \cite{das2021sketchode}, consisting of \(\sim\)10K samples as a simpler dataset to demonstrate complex downstream applications like generation and interpolation.

\subsection{SketchINR: A Compact High-Fidelity Codec}
\begin{figure}
    \centering
    \includegraphics[trim={0 0.3cm 0 0.7cm}, width=\linewidth]{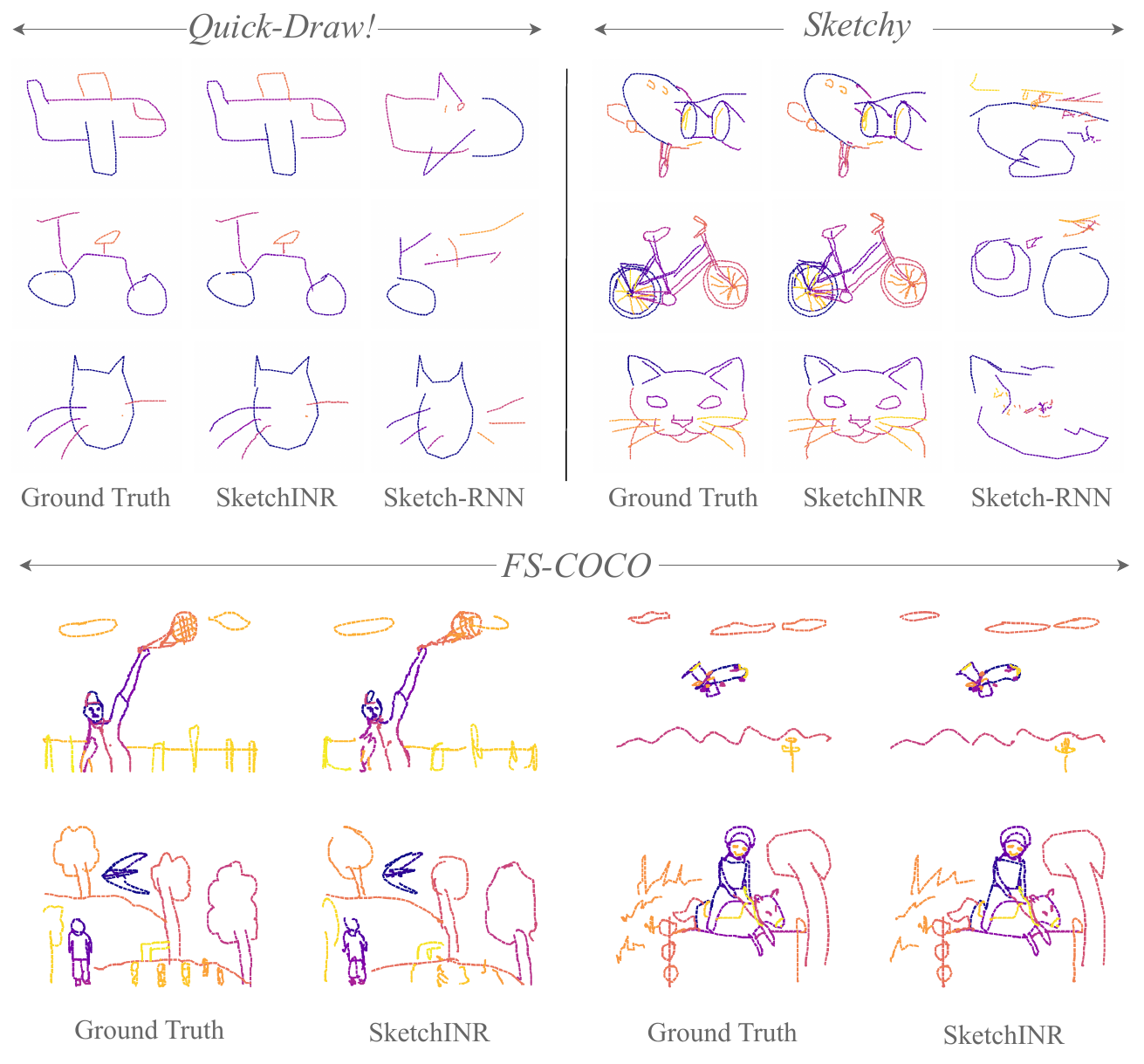}
    \caption{Qualitative reconstruction results for neural sketch representations. SketchINR is uniquely able to scale to sketches of \textit{Sketchy} and \textit{FS-COCO} complexity.\\[-0.7cm]}
    \label{fig: self-recon}
\end{figure}

We begin by demonstrating that SketchINR provides a high fidelity neural representation for sketches. Specifically, we train SketchINR on complex variable length vector sketch datasets such as \textit{FS-COCO} and \textit{Sketchy}. We represent each sketch in these datasets as a fixed size vector with a dataset-specific decoder $f_\theta$. We then render each implicit sketch  \(\nu_i \in \mathcal{V}\) with the same number of strokes as the ground-truth sketch $\mathbf{p}^i$ to evaluate reconstruction quality. 
Following \cite{ha2017neural, das2021sketchode}, we measure reconstruction quality using (i) Chamfer Distance (CD), and (ii) retrieval accuracy (top-10) with a naive sketch retrieval network (ResNet-18) trained on ground-truth sketch vectors $\mathbf{p}$ and evaluated on rendered $\mathbf{\hat{p}}$.
The quantitative results in \cref{tab: self-recon}, show that SketchINR provides substantially higher fidelity encoding than alternative representations like RNN \cite{ha2017neural}, B\'ezier strokes \cite{das2020beziersketch}, ODE dynamics \cite{das2021sketchode}, and stroke embeddings \cite{aksan2020cose} particularly for extremely complex sketches in \textit{FS-COCO}, where SketchRNN and CoSE fail entirely. Qualitative results in \cref{fig: self-recon} show SketchINR to reconstruct simple and complex sketches with high fidelity while SketchRNN performs poorly for \textit{Sketchy} and fails completely on \textit{FS-COCO}.

{
\setlength{\tabcolsep}{5pt}
\begin{table}
    \centering
    \scriptsize
    \begin{tabular}{l c c c c c c c}
        \toprule
         & \multicolumn{3}{c}{FS-COCO \cite{chowdhury2022fs}} & \multicolumn{3}{c}{Sketchy \cite{sangkloy2016sketchy}} \\
         & CD & R@10  & Time (s) & CD & R@10 & Time(s)\\ \midrule
        SketchRNN \cite{ha2017neural}  & -- & -- & -- & 0.16\phantom{0} & 41.65  & 0.11\phantom{0} \\
        CoSE \cite{aksan2020cose} & -- & -- & -- & 0.14\phantom{0} & 45.82 & 0.10\phantom{0} \\
        SketchODE \cite{das2021sketchode}  & 0.41\phantom{0} & 5.17\phantom{0} & 1.68 & 0.17\phantom{0} & 37.20 & 0.13\phantom{0} \\
        B\'ezierSketch \cite{das2020beziersketch} & 0.57\phantom{0} & 2.96\phantom{0} & 1.22 & 0.23\phantom{0}  & 25.72 & 0.08\phantom{0} \\
        \midrule
        \textbf{Ours}   & 0.011 & 15.61 & 0.3\phantom{0} & 0.008 & 57.29 & 0.0007\\ \bottomrule
    \end{tabular}
    \vspace{-0.1cm}
    \caption{Evaluating neural sketch representations' reconstruction quality (Chamfer distance $\downarrow$; recall@10 $\uparrow$) and decoding speed.\\[-0.3cm]}
    \label{tab: self-recon}
\end{table}
}

\begin{figure}[!h]
    \includegraphics[width = \linewidth]{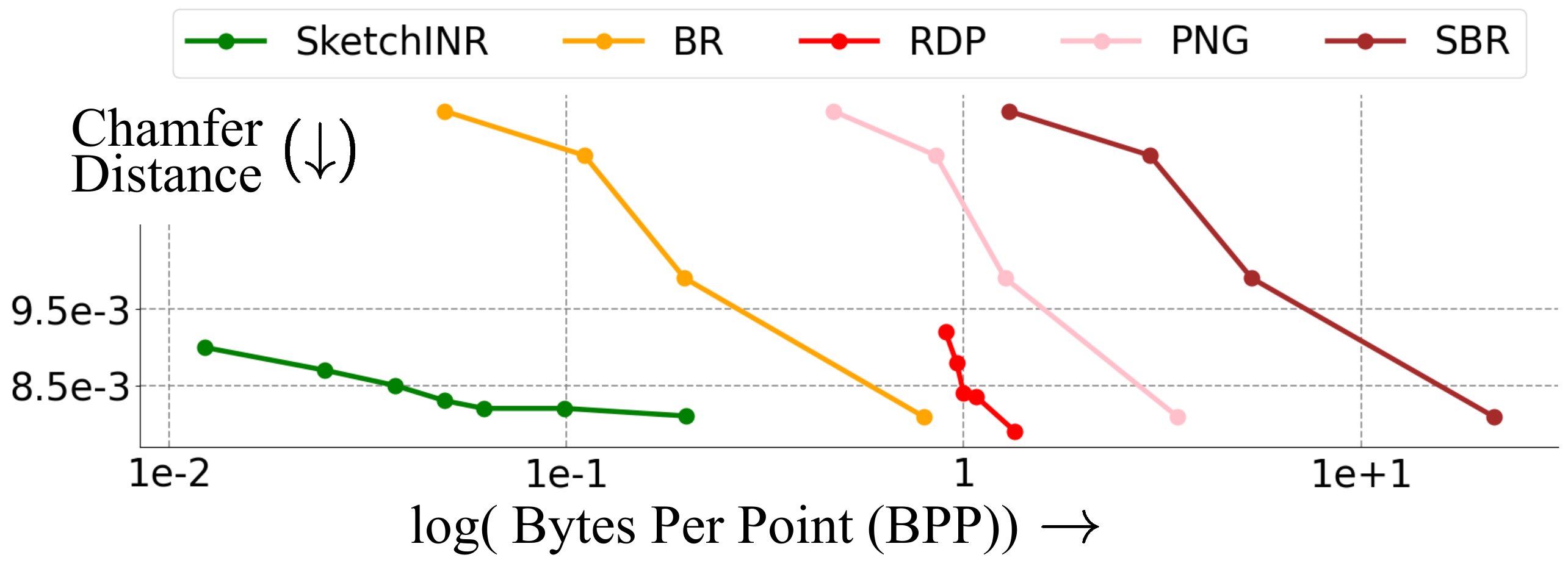}
    \vspace{-6mm}
  \caption{Rate Distortion: SketchINR can encode complex scene sketches from \textit{FS-COCO} \cite{chowdhury2022fs} in highly compact ($\mathbb{R}^{64}$) latent codes. Specifically, despite nearly identical sketch quality (low CD represents higher fidelity), SketchINR has $\sim$$60\times$ lower BPP than PNG raster sketches and $\sim$$10\times$ lower than vectors sketches.\\[-0.3cm]}
  \label{fig: compression}
\end{figure}

\noindent \textbf{Sketch Compression:} We next show that SketchINR's high-fidelity encoding is extremely compact, thus providing an efficient sketch codec. 
INRs store information implicitly \cite{INRcompression} in the network weights ($\theta$) and latent codes ($\nu_i$). To store or transmit a sketch dataset, we only stream weights $\theta$ once (fixed cost) and a compressed latent vector $\nu_i \in \mathbb{R}^{D}$ for each sketch (variable cost). Hence, the space complexity for $N_{D}$ sketches with SketchINR is $|\theta| + N_{D} \times |\nu_i|$ bytes, where $\nu_i$ is $64$ dimensional latent. For streaming a large number of sketches, we discount \(\theta\) as negligible streaming overhead. Comparing to alternative representations, streaming $N_{D}$ (i) binary raster (BR) sketches of size $(N_x \times N_y)$ take $N_{D} \times N_x \times N_y$ bits , and (ii) vector scene sketches with $N$ coordinates and pen states ($\mathbb{R}^{2+1}$) take $N_{D} \times N \times 2 \times 3$ bytes in $16$-bit floating precision ($N \approx 3000$ in \textit{FS-COCO} \cite{chowdhury2022fs}). Raster representations can be further stored by only storing the coordinates ($\mathbb{R}^2$) of pixels containing the sketch (\textit{i.e.}, inked pixels) as sparse binary rasters (SBR). \cref{fig: compression} shows rate-distortion for storing $10,000$ scene sketches \cite{chowdhury2022fs} with $\nu_i$ varying from $128$ to $8$ dimensional latent, and quality for raster and vector forms dropped with downsampling and RDP simplification \cite{douglas1973algorithms} respectively. Quality metrics like PSNR and SSIM are more suited to photos. We use Chamfer Distance as a quality metric for vector sketch {and observe $\sim$$10\times$ and $\sim$$60\times$ more efficiency than vector and raster sketches, respectively}. 
Future works can further optimise our models and thus reduce storage requirements via model compression \cite{HMQ, hawq, mixedprecision2020} and quantisation \cite{aimet}.

\begin{figure}
    \centering
    \includegraphics[trim={0 0.2cm 0 0}, width=\linewidth]{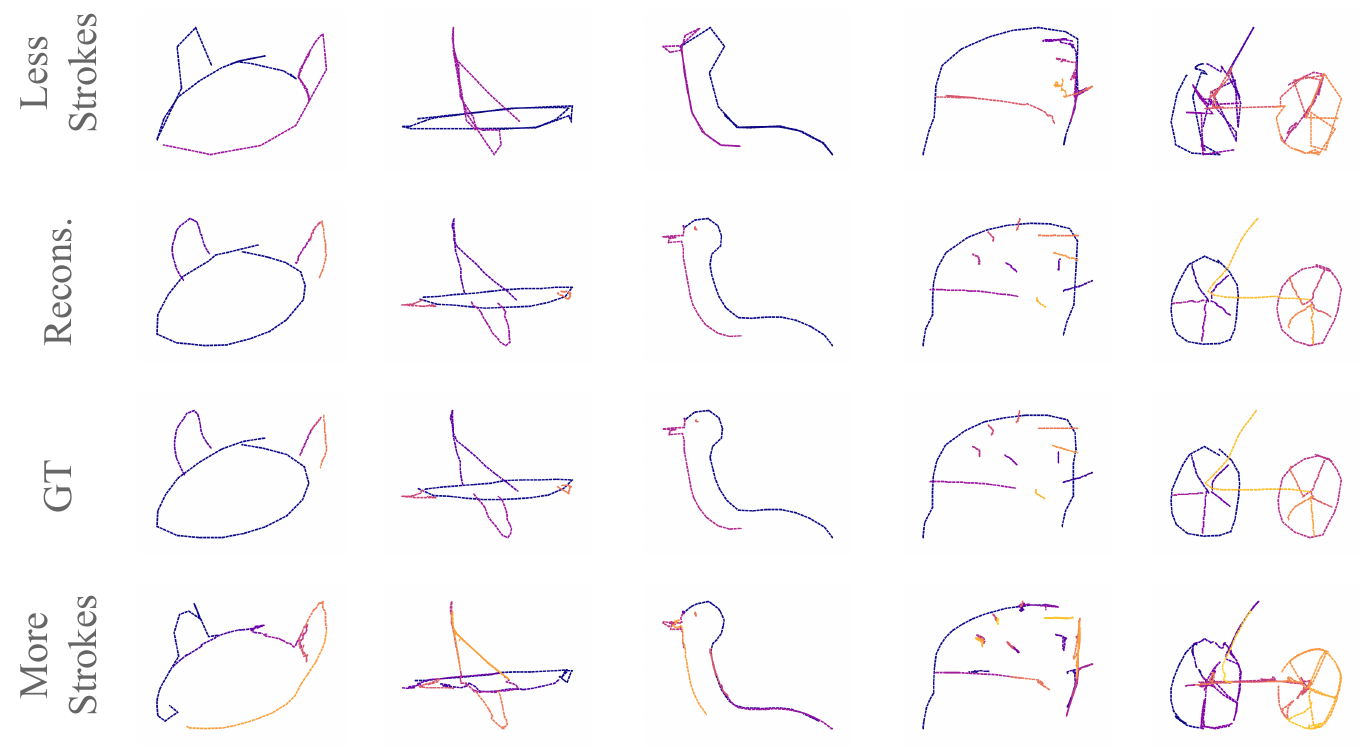}
    \caption{Given a latent code $\nu_i$, we can reconstruct a sketch using learned \(f_\theta\) at multiple resolutions by controlling the number of strokes \(K\) during sampling.\\[-0.7cm]}
    \label{fig: intra-sketch-interpolation}
\end{figure}

\noindent \textbf{Intra-Sketch Variation:} {Unlike prior representations (RNN \cite{ha2017neural}, ODE dynamics \cite{das2021sketchode}), humans do not draw exact stroke-level replica of a raster sketch. Instead, we choose the level of detail, the number of points, and the strokes used to convey the underlying shape. Given a latent code $\nu_i$, \textit{SketchINR} allows users to explicitly choose $N$ and $K$, where the number of points $p = \{p_{1}, \dots, p_{N}\}$ and strokes $s = \{s_{1}, \dots, s_{K}\}$. Choosing a low $N$ and $K$ leads to abstract self-reconstructions with fewer points and strokes. \cref{fig: intra-sketch-interpolation}(top to bottom) shows that we can control $K$  to decrease or increase the level detail in the sketch.}

\subsection{Latent Space Interpolation (Creative Mixing)} 
\begin{figure}
    \centering
    \includegraphics[trim={1.5cm 0.3cm 1cm 0}, width=0.9\linewidth]{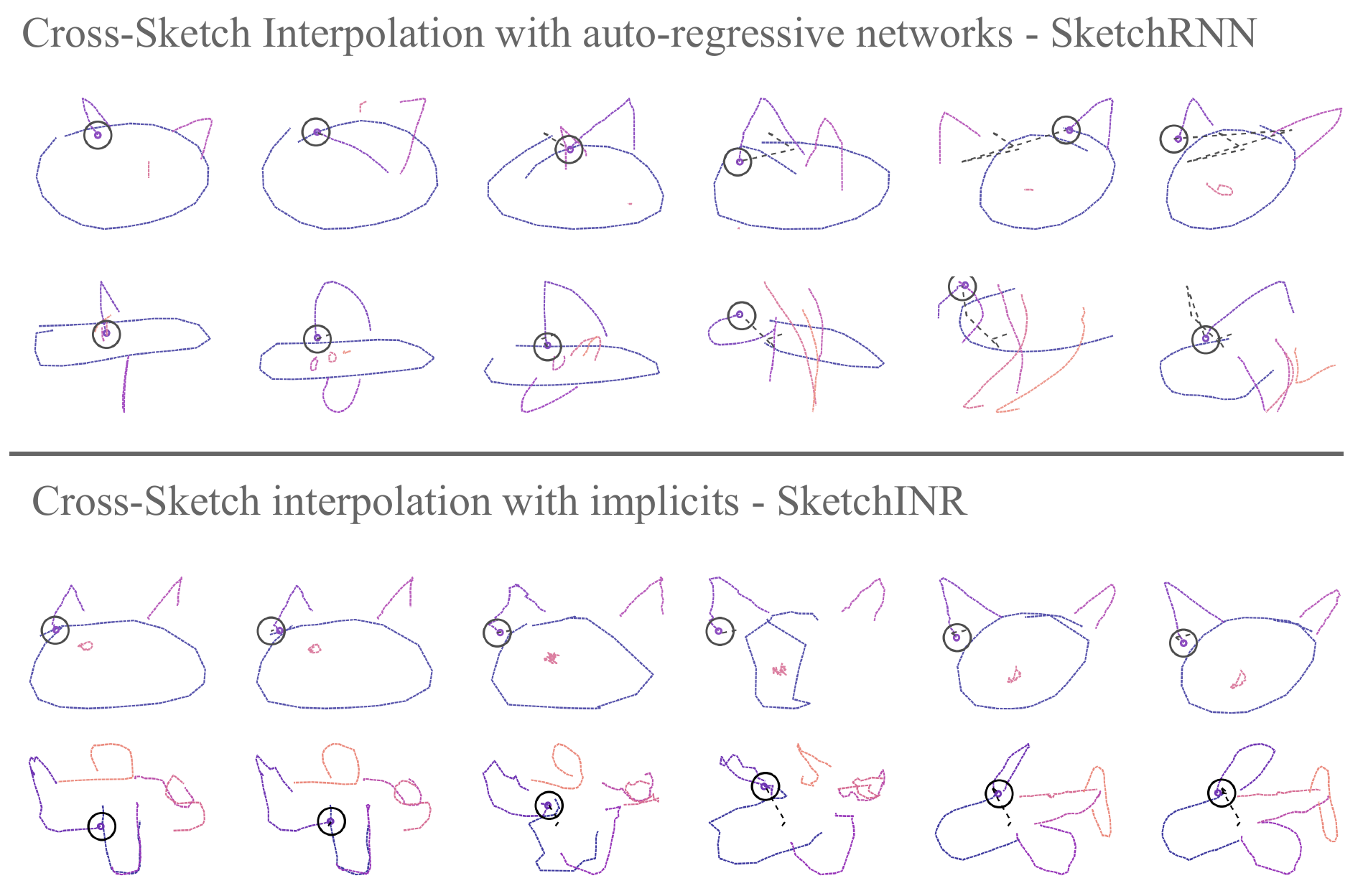}
    \caption{Latent space interpolation shows continuity and translation from one sketch concept to another.\\[-0.7cm]}
    \label{fig: latent-interpolation}
\end{figure}

Interpolation of learned latents allows us to explore latent space continuity. Auto-regressive modelling of sketches builds poor latents as explored by Das \textit{et. al.} \cite{das2023chirodiff} due to the lack of a full visibility of the sketch at any decoding timestep. Specifically, auto-regressive models use the partially predicted sketch along with the sketch's latent representation to predict new way-points with consistency. This, however, introduces noise in the latent \(\rightarrow\) sketch decoding so that \emph{small changes in these latents lead to big changes in the final sketch}. Deterministic latent to sketch decoding solves this problem as noise is not added at any point, leading to meaningful sketch interpolation from corresponding latent space interpolations. We visualise in \cref{fig: latent-interpolation} how one sketch morphs into another by reconstructing from $f_{\theta}(\nu', t_{n}, \Delta t, s_{k}, \Delta s)$, where $\nu'$ is sampled along a \textit{latent walk}: $\nu' = (1 - \delta) \nu_{1} + \delta \nu_{2}$, for $\delta \in [0, 1]$. Interpolating between two sketches in the continuous latent space gives creative mixing -- modifying the sketch of airplane-1 (facing right) into airplane-2 (facing left) by varying $\delta$. \cref{fig: latent-interpolation} shows that small changes in the latent (columns) leads to smoother changes in sketch space for SketchINR. Meanwhile SketchRNN can exhibit big jumps that can zig-zag over successive latent increments, as indicated by an example tracked point in the figure. 

\subsection{INR inversion for sketch completion}
\begin{figure}
    \centering
    \includegraphics[trim={0 1.5cm 0 1cm}, width=\linewidth]{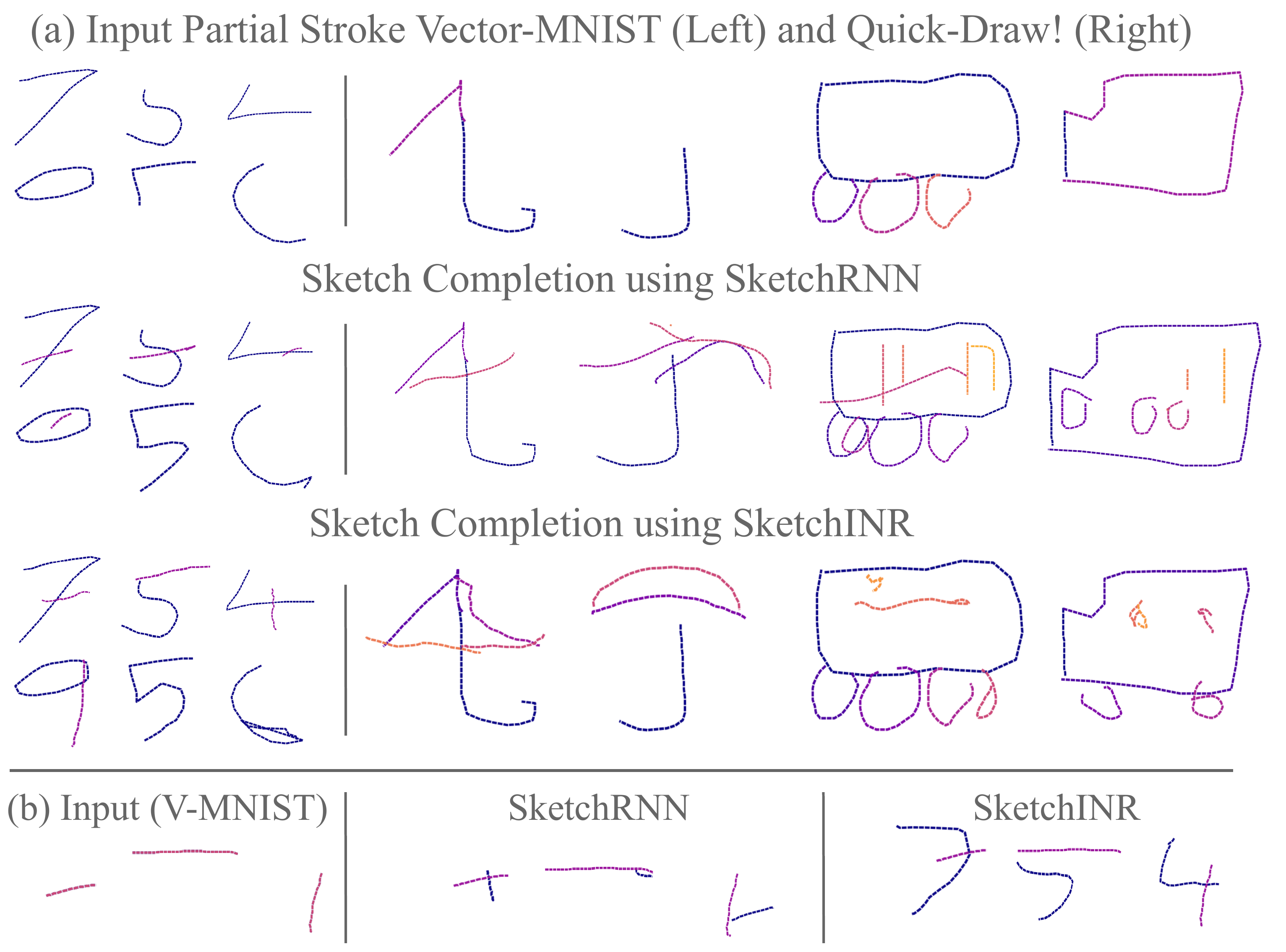}
    \caption{We perform sketch completion by inverting learned implicit representations and decoding for occluded strokes or way-points. (a) Temporal completion (b) Unlike auto-regressive methods like SketchRNN \cite{ha2017neural}, we can perform a-temporal (or inverse-temporal) completion, where given the second half of a partial sketch, we can successfully reconstruct the first part.\\[-0.8cm]}
    \label{fig:completion}
\end{figure}

Neural networks used to approximate implicit representations have been inverted in parallel literature \cite{hertz2022spaghetti} to appropriate samples not seen during training. This inversion,  to arrive at a latent \(\nu\) that describes the new sample, is usually made possible by learned priors in \(f_\theta\). Specifically, an implicit decoder \(f_\theta\) is frozen with latent \(\nu\) optimised for \(f_\theta(\nu,\dots)\) against the ground truth. This behaviour is easily replicated in SketchINR, where we optimise the latent \(\nu\) to describe sketch samples (\cref{fig: self-recon}). To implement sketch completion, we obtain a sketch descriptor \(\nu\) from a partial sketch by just optimising for specific time-steps (\eg 1/4th of a sketch; \(t \in [0,\frac{1}{4})\)) available in a partial doodle. Then, by sampling for rest of the timesteps we can complete the incomplete sketch. We summarise some qualitative sketch completion results in \cref{fig:completion}(a). Completed sketches are varied because of randomness in optimisation, as partial sketches are often ambiguous. INR inversion thus provides a novel approach to sketch completion, previously dominated by auto-regressive and models. Unlike auto-regressive models SketchINR can also perform \emph{a-temporal} completion, inferring the first half of a sketch given the second half of the sketch (\cref{fig:completion}b). We remark that popular INR applications such as super-resolution \cite{nguyen2023single} and point-cloud densification \cite{park2019deepsdf} perform interpolation after inversion on sparse inputs (\eg: $t \in [0,1)$ given $t\in\{0,\frac{1}{4},\frac{1}{2},\frac{3}{4}\}$), while sketch completion requires extrapolation (\eg: $t \in [0,1)$ given $t\in [0,\frac{1}{4})$), making it more analogous to visual outpainting. 

\subsection{Sketch Generation}
We can perform conditional and unconditional sequential vector sketch generation after learning a generative model for sketch latents as described in \cref{sec:multi}.

\begin{figure}
    \centering
    \includegraphics[width=\linewidth]{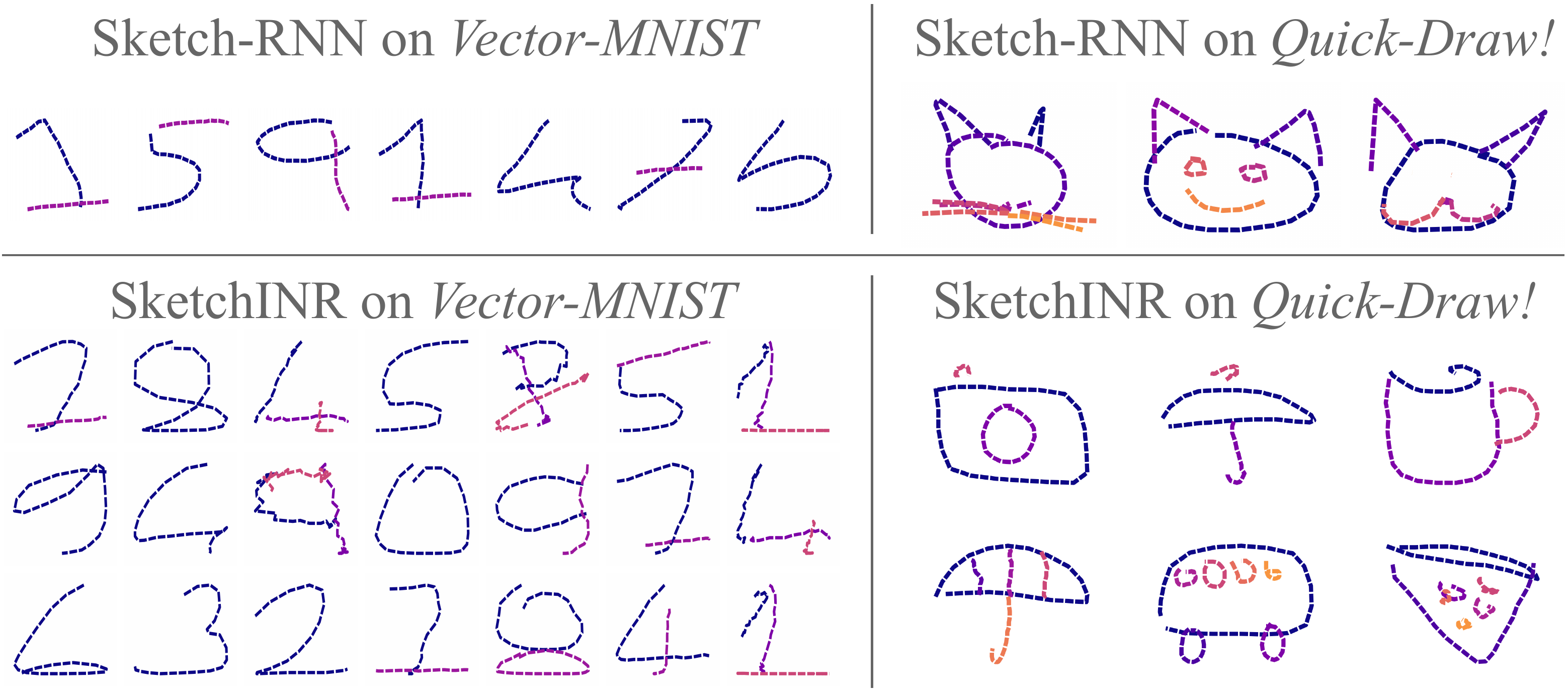}
    \caption{Unconditional sketch generation. Sampling sketch latents $\nu$ as well as stroke and point complexity $(K,T)$.\\[-0.7cm]}
    \label{fig: unconditional-generation}
\end{figure}

\noindent\textbf{Unconditional Generation: }
We unconditionally generate sketches by random sampling \(\epsilon \sim \mathcal{N}(\textbf{0},\textbf{I})\) with the learnt VAE latent space, obtaining novel sketch descriptors \(\nu_i \in \mathcal{V}\).  We further sample \(f_\theta (\nu_i, \dots) \) with varying timestamps $T \in [100, 300]$ and strokestamps $K \in [10, 30]$ obtaining sketches at varying levels of abstraction. Qualitative results in \cref{fig: unconditional-generation} show diverse and plausible samples from \textit{Vector-MNIST} \cite{das2021sketchode} and \textit{Quick-Draw!} \cite{ha2017neural} datasets. 

\begin{figure}[!h]
    \centering   
    \includegraphics[trim={0 0.5cm 0 1cm}, width=\linewidth]{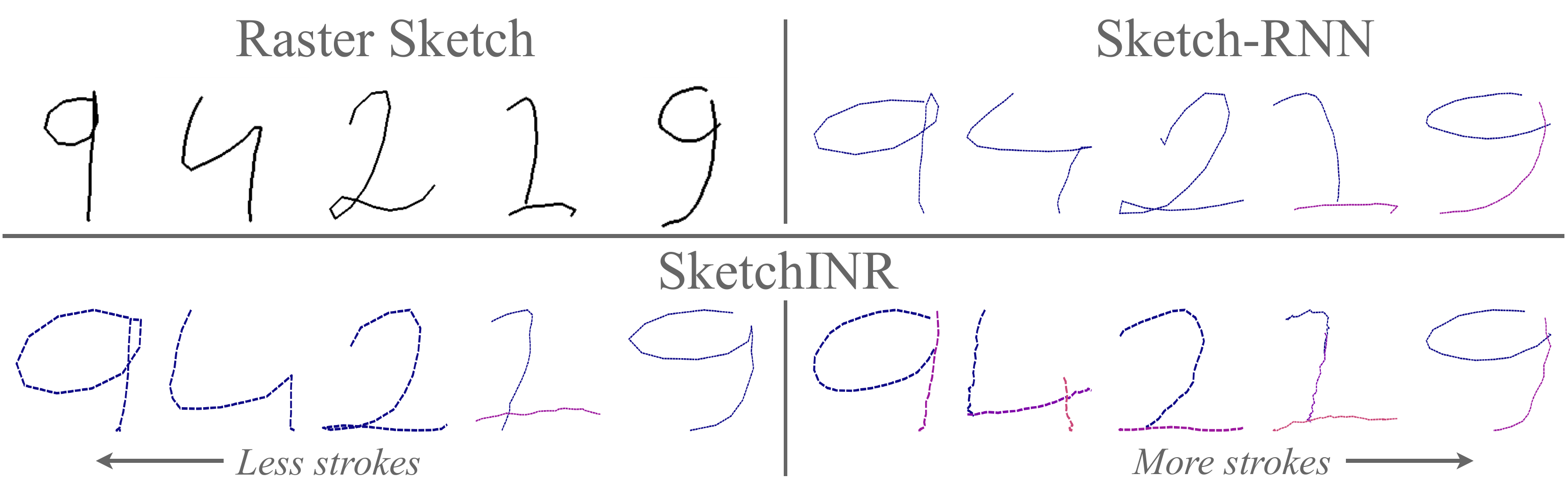}
{
  \caption{Conditional sketch generation (vectorisation) on \textit{Vector-MNIST}. Raster sketches (top left) can have multiple vector forms, \eg, sketched with different numbers of strokes (bottom).\\[-0.7cm]}%
  \label{fig: conditional-generation}
}
\end{figure}

\noindent\textbf{Conditional Generation: }
To condition sketch generation, we use our pre-trained VAE encoder \(E\) to embed raster sketches, which can then be vectorised by the decoder $f_\theta$. \cref{fig: conditional-generation} shows qualitative results for sequential vector sketch generation from raster images, where we obtain comparable results with auto-regressive generative modelling \cite{ha2017neural}. We observe that treating vectorisation as a generative problem rather than a deterministic raster-to-vector mapping \cite{bhunia2021vectorization} leads us to have variations in generated vectors particularly in the form of stroke order resembling real copies of sketches made by humans. 

\subsection{Ablations and Discussion}

\begin{figure}
    \centering
    \includegraphics[trim={0 0.7cm 0 0}, width=\linewidth]{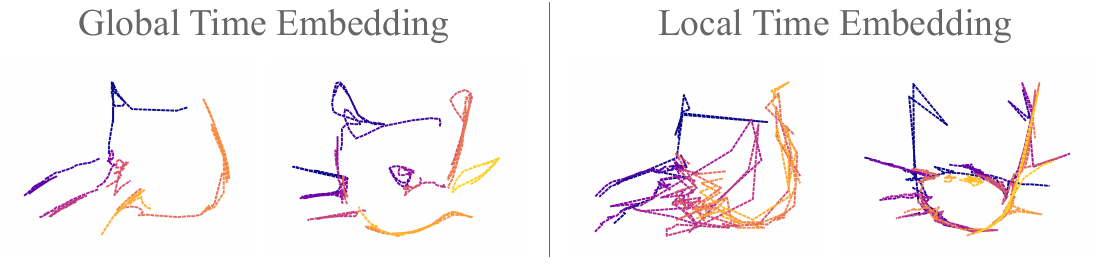}
    \caption{Global vs Local time embedding: Global embedding gives \textit{smooth} sketches compared to local, which gives jittering.\\[-0.8cm]}
    \label{fig:local-vs-global-time}
\end{figure}

\noindent \textbf{Global vs Local Time Modelling:} We model \(J\) timestamps across \(K\) strokes, by distributing timestamps uniformly over each stroke such that each stroke has \(\frac{J}{K}\) timestamps. In other words, time ($t, \Delta t$) is modelled \textit{globally} for the entire sketch $\mathbf{p} = \{ \mathrm{p}_{1}, \dots, \mathrm{p}_{N} \}$. An alternative is to model ($t, \Delta t$) as \(0 \text{to} 1\) for each stroke separately (local time-modelling) instead of the entire sketch. Particularly, each of the $K$ strokes are modelled using $N$ points, representing the entire sketch using $N \cdot K$ points as $\mathbf{p} = \{ \mathrm{p}_{1}, \dots, \mathrm{p}_{N\cdot K} \}$. Visually inspecting global vs local time embedding shows that local modelling leads to \textit{jittery} sketches (see \cref{fig:local-vs-global-time}), whereas global modelling gives \textit{visually smooth} sketches. We hypothesise that time is a global property of a sketch and not an independent property localised to each stroke. Hence, modelling time locally puts an additional optimisation burden on our implicit decoder to ensure smoothness and continuity, resulting in poorly reconstructed sketches.

\noindent \textbf{Fixed vs Variable Smoothing Factor ($\gamma$):} The intensity map $\mathrm{I}_{k}$ in \cref{eq: dist_field} guides our raster-based visual loss $\mathcal{L}_{V}$. A key component in $\mathrm{I}_{k}$ is the smoothing parameter $\gamma$ -- low $\gamma$ (thicker $\mathrm{I}_{k}$) gives less accurate but smoother values on the entire visual region (easier training); high $\gamma$ (thinner $\mathrm{I}_{k}$) gives more accurate but sharper (steeper) intensity-gradients near ground-truth strokes (harder training). While we use a fixed \(\gamma\) with MSE Loss in \cref{equation: all} to balance out the optimisation complexity, we note that an alternative is to vary \(\gamma\) every fixed number of steps with a scheduler to allow faster optimisation. Specifically, we vary \(\gamma\) as \(20 \text{to} 200\) on a linear scale by incremental \(\Delta \gamma\) every few iterations. Despite its theoretical advantage \cite{LM-algorithm, gauss-newton-approx}, our initial experiments suggest only minor improvement that reduces training time by $6.1\%$. A detailed study of bundle-adjustment \cite{LM-algorithm} for visual loss is an interesting future work.

\vspace{0.2cm}
\noindent\textbf{Limitations and Future Work}\quad
Despite it's efficiency as a representation for hand-drawn sketches, SketchINR suffers from a number of limitations. We naturally share limitations with other implicit representations including optimisation-based encoding and poor cross category generalisation. In addition to these, a core limitation lies in slow convergence due to pixel space optimisation. While vector-space loss in the form of mean squared error mitigates this to some extent (\cref{fig: effect-gamma}), convergence is still slow and can possibly be improved by better engineering the implicit function. Finally, some generated sketches have jagged edges, which could be explicitly punished with stroke gradient (slope) based regularisation in future work to optimise for smoother strokes.

\vspace{-0.1cm}
\section{Conclusion}
We introduce visual implicits to represent vector sketches with compressed latent descriptors. This neural representation provides a high-fidelity and compact representation that raises the possibility of a sketch-specific codec for compactly representing large sketch datasets. Our SketchINR can decode sketches at varying levels of detail with controllable number of strokes, and provides superior cross-sketch interpolation between implicits, demonstrating a smoother latent space than auto-regressive models. Applications such as sketch-completion are also supported, including a-temporal completion not available with auto-regressive models. Decoding is inherently parallel and can be over $100\times$ faster than autoregressive models in practice.

{
    \small
    \bibliographystyle{ieeenat_fullname}
    \bibliography{main}
}

\end{document}

%% file: preamble.tex
\usepackage[dvipsnames]{xcolor}

\usepackage{pgfplots}
\pgfplotsset{compat=1.18,
    /pgfplots/ybar legend/.style={
    /pgfplots/legend image code/.code={%
       \draw[##1,/tikz/.cd,yshift=-0.25em]
        (0cm,0cm) rectangle (3pt,0.8em);},
   },
}
\usepackage{tikz}

\newcommand{\cut}[1]{} 
\usepackage{floatrow}
\newfloatcommand{capbtabbox}{table}[][\FBwidth]
\usepackage{mathtools}
\usepackage{amsthm}
\usepackage{blindtext}
\usepackage{makecell}
\usepackage{dsfont}
\usepackage{arydshln}

\definecolor{gamma_high}{HTML}{ED7D58} 
\definecolor{gamma_low}{HTML}{0D0887}

\definecolor{deepGreen}{RGB}{0,153,0}
\definecolor{orange}{RGB}{255,125,0}
\definecolor{Gray}{gray}{0.9}

\makeatletter
\apptocmd\@maketitle{{\myfigure{}\par}}{}{}
\makeatother
\usepackage{capt-of,etoolbox}